\documentclass[runningheads]{llncs}
\usepackage{graphicx}

\usepackage{tikz}
\usepackage{comment}
\usepackage{amsmath,amssymb} 
\usepackage{color}
\usepackage{graphicx}
\usepackage{amsmath}
\usepackage{amssymb}
\usepackage{booktabs}
\usepackage{algorithm}
\usepackage{algpseudocode}
\usepackage{makecell}
\usepackage{xcolor}
\usepackage{t1enc}
\usepackage{adjustbox}
\usepackage{subfig}
\usepackage{tabularx}
\usepackage{multirow}

\usepackage[flushleft]{threeparttable} 
\usepackage{booktabs,caption}

\usepackage[accsupp]{axessibility}  

\begin{document}
\pagestyle{headings}
\mainmatter
\def\ECCVSubNumber{1430}  

\title{Addressing Client Drift in Federated Continual Learning with Adaptive Optimization} 


\titlerunning{Addressing Client Drift in FCL}
%
\author{Yeshwanth Venkatesha \and
Youngeun Kim \and
Hyoungseob Park \and
Yuhang Li \and
Priyadarshini Panda}
\authorrunning{Y. Venkatesha et al.}
%
\institute{Yale University, New Haven CT 06511, USA\\
\email{\{yeshwanth.venkatesha, youngeun.kim, hyoungseob.park, yuhang.li, priya.panda\}@yale.edu}}
\maketitle

\begin{abstract}
Federated learning has been extensively studied and is the prevalent method for privacy-preserving distributed learning in edge devices. Correspondingly, continual learning is an emerging field targeted towards learning multiple tasks sequentially.
However, there is little attention towards additional challenges emerging when federated aggregation is performed in a continual learning system.
We identify \textit{client drift} as one of the key weaknesses that arises when vanilla federated averaging is applied in such a system, especially since each client can independently have different order of tasks.
We outline a framework for performing Federated Continual Learning (FCL) by using NetTailor as a candidate continual learning approach and show the extent of the problem of client drift. We show that adaptive federated optimization can reduce the adverse impact of client drift and showcase its effectiveness on CIFAR100, MiniImagenet, and Decathlon benchmarks. Further, we provide an empirical analysis highlighting the interplay between different hyperparameters such as client and server learning rates, the number of local training iterations, and communication rounds. Finally, we evaluate our framework on useful characteristics of federated learning systems such as scalability, robustness to the skewness in clients' data distribution, and stragglers.
\keywords{Federated Learning, Continual Learning, Client Drift, Adaptive Optimization}
\end{abstract}

\section{Introduction}\label{sec:intro}
Federated learning has drawn widespread attention to utilizing millions of edge devices that accumulate massive amounts of data while preserving data privacy \cite{fl_google,kairouz2019advances,konevcny2016federated,mohri2019agnostic,li2020federated}.
Since these devices constantly generate enormous amounts of data, we need models that adapt to the trends in the new data.
Several studies have addressed this problem of learning on a sequential stream of data termed continual learning or lifelong learning \cite{continual_survey_1,chen2018lifelong}.
While both federated learning and continual learning have been widely studied independently in recent years, the effect of federated aggregation in a continual learning system, referred to as federated continual learning (FCL) has not been sufficiently explored despite its relevance to practical scenarios.



\begin{figure}[t]
\centering

\subfloat[Client Drift. 
]{%
  \includegraphics[clip,width=0.46\textwidth]{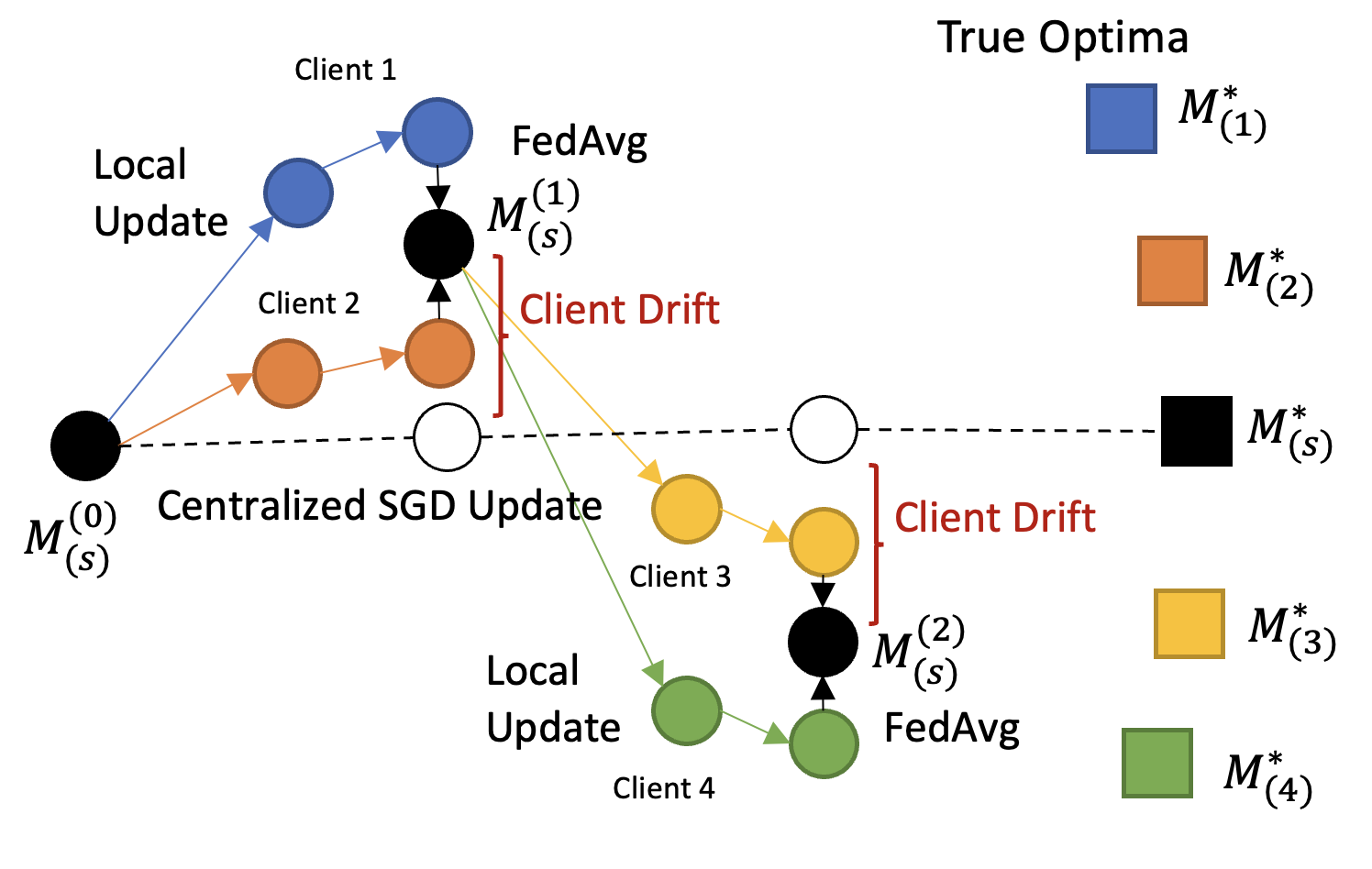}%
  \label{fig:client_drift_revised}
}
\qquad
\subfloat[Relative order of tasks. 
]{%
  \includegraphics[clip,width=0.46\textwidth]{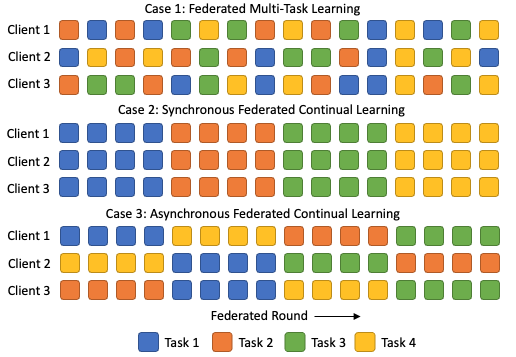}%
  \label{fig:task_order_revised}
}

\caption{Illustrations showing the two main challenges in a federated continual learning system (a) Illustration of Client Drift in a 4 client system. (b) The relative order of tasks in the clients corresponding to three practical cases. }
\vspace{-7mm}
\end{figure}

In an FCL system, the challenges are twofold. First, following the popular formulation of continual learning as a sequence of tasks, the challenge is to learn new tasks while avoiding the overwriting of previously learned tasks, commonly referred to as \textit{Catastrophic Forgetting} \cite{goodfellow2013empirical,ewc}. 
Second, similar to catastrophic forgetting between tasks in continual learning, federated learning handles heterogeneous data distributions from different clients resulting in a phenomenon called \textit{client drift} \cite{scaffold}. As illustrated in Fig. \ref{fig:client_drift_revised}, when the data is heterogeneous, each of the clients pushes the model in a different direction in the optimization space and takes the model away from the true optima resulting in client drift. 
Further, since there are multiple clients, the relative order of tasks in each client determines the overall client drift. We describe three practical cases with respect to the ordering of tasks. First, when all the clients contain the data for all the tasks simultaneously, defining the scenario of federated multi-task learning (FMTL) which has been explored previously \cite{MOCHA,dinh2021fedu,marfoq2021federated}. Second, the ideal case of federated continual learning where all the clients synchronously train on the same sequence of tasks. Finally, the case where each client independently trains on a different sequence of tasks. Fig. \ref{fig:task_order_revised} visually depicts the aforementioned order of tasks. Note that for simplicity we assume every client contains the dataset for all the tasks while in practice it is possible for some clients to not have some of the tasks. 
Depending on the order of tasks, the system will have different trade-offs in terms of client drift and catastrophic forgetting. Intuitively, since the data from all the tasks are repeated across all clients over several rounds of training, the case of FMTL will have the best performance in terms of both client drift and catastrophic forgetting given a model with sufficient capacity. In the case of synchronous FCL, the problems of catastrophic forgetting and client drift are akin to that of standard continual learning and federated learning respectively. On the other hand, in asynchronous FCL the later rounds can overwrite the knowledge accumulated by the clients in the former rounds aggravating the problem of client drift.

In this paper, we focus on the problem of additional client drift in asynchronous FCL. We use a dynamic architecture approach based on NetTailor \cite{nettailor} which has independent modules for each task thereby eliminating the problem of catastrophic forgetting and isolating the effect of client drift.
We show that using adaptive federated optimization \cite{FedOpt} can tackle this problem by showing its effectiveness by visualizing the client drift between models from consecutive rounds. 
We provide a comprehensive analysis of our system spanning different hyperparameters such as the number of communication rounds and learning rates. Further, we evaluate the performance with respect to challenges in a large-scale federated system such as scalability, data distribution, and stragglers.
Our contributions are summarized as follows:
\begin{itemize}
\itemsep0em 
    \item We examine the problem of federated continual learning from the perspective of relative task order in the clients and identify client drift as the central challenge. We develop a framework for FCL based on the NetTailor approach to show the extent of client drift.
    \item We show that adaptive federated optimization (FedOpt) reduces the effect of client drift. With FEDADAM, we achieve an improvement of 2.45\% in terms of average accuracy while reducing the client drift by 4.92\% over standard FedAvg for a 5 client system trained on CIFAR100 split into 10 tasks.
    \item We provide a comprehensive view of the hyperparameter space by performing ablations studies on client and server learning rates, number of communication rounds, and local epochs. We analyze the robustness of the system to different challenges of federated learning such as non-IID distribution, stragglers, and scalability. 
\end{itemize}

\section{Related work}
The popular continual learning methods can be broadly classified into three categories --- (1) Regularization approaches (2) Dynamic Architecture approaches (3) Memory Replay and Complementary Learning System approaches \cite{continual_survey_1,continual_survey_2}. Regularization-based approaches train a single model by using some form of regularization while learning new tasks so as to reduce the overwriting of the model parameters learned from the previous tasks \cite{lwf,jung2016less,azizpour2014cnn,donahue2014decaf,ewc,CL_SI,maltoni2019continuous}.
Dynamic architecture approaches modify the network architecture according to the task either by adding additional modules or by using a subset of modules \cite{progressive_neural_networks,DEN,sarwar2019incremental,li2019learn,xu2018reinforced,yoon2019scalable,zhou2012online,xiao2014error,draelos2017neurogenesis,part2016incremental,part2017incremental,nettailor}.
Memory replay methods store a subset of the data from the previous tasks or synthesize using a generative model and use it to retrain along with the new tasks and avoid the drop in performance on previous tasks \cite{french1997pseudo,rebuffi2017icarl,lopez2017gradient,chaudhry2018efficient,riemer2018learning,shin2017continual,guo2019improved,chaudhry2019continual}. Complementary learning systems
use different modules to learn and store generic long-term knowledge and task-specific short-term knowledge \cite{robins1995catastrophic,soltoggio2015short,hinton1987using}.

While regularization approaches are generally more efficient as they use a single model for all the tasks, training such methods across multiple clients in a federated learning paradigm involves significant interference both between tasks and data distribution in clients. On the other hand, since memory replay methods require storing a subset of past data to train the model on new tasks, it is not viable to retain data privacy while training such systems in a federated paradigm.
Although dynamic architecture approaches require additional memory as we add more modules to the model architecture for each new task, they offer flexibility in terms of adding more tasks to each client independently.   
Moreover, the dynamic architecture approach decouples the orthogonal problems of catastrophic forgetting and client drift resulting in a simpler design. 
NetTailor \cite{nettailor} is one of the state-of-the-art continual learning methods based on the dynamic architecture approach which uses a common backbone network and adds additional lightweight task-specific branches resulting in a simple framework to implement continual learning without catastrophic forgetting. This provides a natural way of aggregating only the task-specific parameters to share knowledge between clients and hence is easily scalable to a large number of tasks. This further reduces the communication cost since only the gradients of task specific lightweight parameters are shared.
We use NetTailor \cite{nettailor} as the candidate approach to show our findings. However, our framework can be extended to other continual learning methods as well.

Vanilla federated averaging (FedAvg) \cite{fl_google} suffers a performance drop when the data is not distributed in an independent and identical (non-IID) fashion among the clients \cite{zhao2018federated}. 
There have been multiple works to improve federated averaging such as FedProx \cite{FedProx}, which uses a regularization term to stabilize federated training, and FedMA \cite{FedMA}, which performs a layer-wise matching and averaging of the models to handle heterogeneous data. The authors of SCAFFOLD \cite{scaffold} use control variates to counter the effect of client drift and stabilize the training process. Further, the authors of Adaptive Federated Optimization (FedOpt) \cite{FedOpt} generalize the FedAvg and apply adaptive optimization algorithms such as ADAM and ADAGRAD in the federated aggregation process. This adaptive optimization makes the federated aggregation more smooth as a result provides an effective way to avoid client drift in federated continual learning systems.

To the best of our knowledge, FedWeIT \cite{fedweit} is the only work that attempts to address the problem of federated continual learning. FedWeIT uses weighted inter-client knowledge transfer to share constructive knowledge between clients having related tasks and avoid interference between clients having unrelated tasks. It achieves this by decomposing the network weights into global federated parameters and sparse task-specific parameters and the knowledge sharing is made selective by sharing a weighted combination of their task-specific parameters. We take an alternative approach of dynamic architecture with disjoint task-specific parameters thereby removing task interference altogether and at the same time reducing the overall communication cost. While FedWeIT provides a more efficient model with shared weights, our framework achieves higher accuracy since the interference between different task parameters is eliminated as well as effect of client drift is reduced with FEDADAM. We provide a comparison with FedWeIT on CIFAR100 dataset in section \ref{section:comp_prev_work}. Further, we provide experiments on the Decathlon benchmark which is a well-known benchmark for evaluating continual learning algorithms as compared to the simulated Non-IID 50 dataset used in FedWeIT. Additionally, we provide an in-depth analysis of different hyperparameters and challenges of federated learning such as scalability, non-IID distribution, robustness to stragglers.

\section{Preliminaries}

\vspace{-3mm} \subsection{Federated Learning} \vspace{-2mm}

A standard federated learning system consists of a set of $N$ clients and a central server. The central server broadcasts the initial model parameters (say $M_{(s)}^{(0)}$) to the clients to initialize the training process. 
Each of the clients $c = 1, 2, ..., N$, store their own private data $D^{(c)}$ which is used to obtain locally trained model (say $M_{(c)}$).
The clients accumulated gradients from local training and periodically communicate them to the central server for aggregation. Each iteration of communication of the gradients from clients to the server followed by broadcasting of the updated model back to the clients is defined as a federated learning \textit{round}. At \textit{round} $r$, the gradients $\Delta M_{(c)}^{(r)}$  from a subset of participating clients $S^{r} (\ll N)$ is sent to the central server. 
The central server performs a weighted average of the gradients to obtain the updated global model as shown in Eq. \ref{fl_equation}.
\begin{equation}\label{fl_equation}
M^{(r + 1)}_{(s)} = M^{(r)}_{(s)} + \frac{1}{\sum_{c \epsilon S^{(r)}} |D_{(c)}^{(r)}|}\sum_{c \epsilon S^{(r)}}|D_{(c)}^{(r)}|\Delta M_{(c)}^{(r)},
\end{equation}
where $|D^{(c)}_{r}|$ denotes the number of data samples used for local training in client $c$ at round $r$. This aggregation of the model updates from the clients is described as FedAvg algorithm in \cite{fl_google}. 

\vspace{-3mm} \subsection{Client Drift} \vspace{-2mm}
The major problem with federated learning systems is that of client drift \cite{scaffold} where the updates from one client overwrites the model weights learned with data from other clients. 
This can be estimated by observing the distance from the initial model and the trained model at a client. Formally, client drift at round $r$ is defined as the expected amount by which the client model deviates from the starting model received from the server,
\begin{equation}
    \mathcal{E}^{(r)} = \frac{1}{KN} \sum_{k = 1}^{K} \sum_{c = 1}^{N} \mathbb{E}[||M^{(r)}_{(c, k)} - M_s^{(r - 1)}||^2].
    \label{eq:client-drift}
\end{equation}
Here, $c \in [N]$ denotes the clients, and $k \in [K]$ denotes the local training epochs over the client's dataset. $||\cdot||$ is a distance metric such as Euclidean or Cosine distance defined over the model weights.

\vspace{-3mm} \subsection{NetTailor} \vspace{-2mm}
The architecture of NetTailor \cite{nettailor} involves a common backbone network trained on a large dataset such as ImageNet and the task-specific blocks interspersed between the layers. Formally, a standard convolutional network implements the following function
\begin{equation}
    f(x) = (G_L \circ G_{L-1} \circ \cdots \circ G_{1})(x),
\end{equation}
where $G_l$ is a function at layer $l$ such as convolution, spatial pooling, normalization, etc. NetTailor uses this as a backbone and augments it with a set of task-specific proxy layers $\{P_{t}^{(l',l)}(\cdot)\}_{l' = 1}^{l-1}$ that introduce skip connections between layer $l'$ and layer $l$ for each task $t$. These proxy branches are computationally lightweight consisting of max pooling and 1x1 convolution layers. For example, compared to standard ResNet blocks which contain two 3x3 convolution layers, these contain $\frac{1}{18}$ of parameters and perform $\frac{1}{18}$ of operations. In practice, it is often redundant to have proxy layers connecting all the previous layers. Hence, the number of proxy layers can be limited to the previous $k$ layers. The activation of each layer is computed by taking a linear combination of the backbone layer and the proxy layers. 
Formally, the activations at layer $l$ for task $t$ can be expressed as, 
\begin{equation}
    x_l = \alpha_l^lG_l(x_{l - 1}) + \sum_{l' = max(l - k, 1)}^{l - 1}\alpha^l_{l'} P_{t}^{(l', l)}(x_{l'}).
\end{equation}
Here $\{\alpha_{l'}^l\}^l_{l'= max(l - k, 1)} \in [0,1]$ are the set of parameters denoting the importance of the branches that are used to enable or disable the branches. Since all the branches are not necessary for all tasks, this architecture is further pruned according to the complexity of tasks. The fact that NetTailor keeps the backbone network frozen and only updates the proxy branches aids in reducing the communication cost in federated learning since we only have to communicate the task-specific parameters and their gradients. 

\section{Federated Continual Learning Framework}\label{section:framework}
\begin{figure*}[t!]
  \begin{center}
    \includegraphics[width=0.8\textwidth]{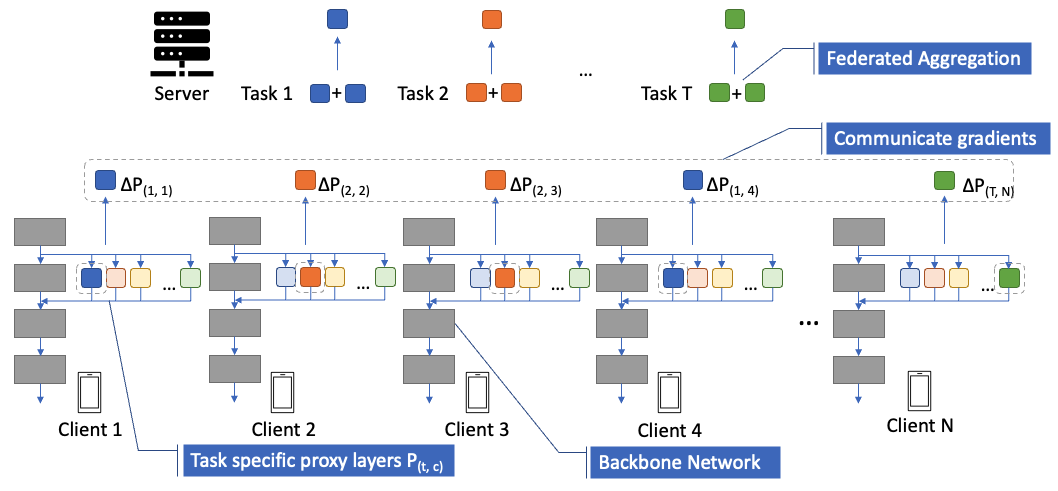}
  \end{center}
  \caption{
     Illustration representing our overall approach. The model in each client comprises a backbone network and a set of task-specific proxy layers for each task. The backbone network is pretrained using a large dataset like ImageNet hence contains the generic knowledge and is kept frozen. Each client independently trains only the task-specific blocks using the local data available. In every communication round, the gradients of the proxy layers are communicated to the server which aggregates these proxy layers and broadcasts them back to all the clients.
  }
   \label{fig:main_method}
   \vspace{-5mm}
\end{figure*}

{
The objective of our work is to showcase the challenge of client drift in federated continual learning systems and show that adaptive federated optimizers can address the effect of client drift. The major challenge in a federated continual learning system is the relative order of tasks in the clients. As illustrated in Fig. \ref{fig:task_order_revised}, we define three different ordering of tasks namely,
\begin{itemize}
    \item \textbf{Case 1:} Federated Multi-Task Learning: Shuffled task order with repetition.
    \item \textbf{Case 2:} Synchronous FCL: Common task order. 
    \item \textbf{Case 3:} Asynchronous FCL: Shuffled task order without repetition.  
\end{itemize}
To study the effect of client drift in the above settings, we design a structured FCL framework based on NetTailor.
The model in each client comprises of backbone parameters $G$ and task-specific parameters $P$. 
As shown in Fig. \ref{fig:main_method}, each client contains a frozen backbone network and a set of trainable task-specific blocks.
We use $P^r_{(t, c)}$ to denote the set of task-specific proxy layers of task $t$ in client $c$ at round $r$, and $P^r_{(t, s)}$ to represent the set of corresponding proxy layers of the server.
At round $r$, each client $c$ has the global parameters sent from the server ($P^r_{(t, c, 0)} = P^{r - 1}_{(t, s)}$).
We then train the local proxy layers in each client for $K$ local epochs on the local client data as
\begin{equation}
    P^r_{(t, c, k)} = P^r_{(t, c, k - 1)} - \mu^r_{(t, c, k)}\nabla P^r_{t, c, k - 1},
\end{equation}\label{eq:local_update}
where $\mu^r_{(t, c, k)}$ is the local learning rate at $k$-th local epoch on client $c$ on task $t$ and $\nabla P^r_{t, c, k - 1}$ is the proxy layer's gradient from the previous local epoch $k-1$.}

Once the clients have performed local training, we obtain the model updates $\Delta P^r_{(t, c)}$, i.e. the difference in the initial model received from the server and the final model after training on the local data.
\begin{equation}
    \Delta P^r_{(t, c)} = P^r_{(t, c, K)} - P^{r - 1}_{(t, s)}.
    \label{eq:gradient_calc}
\end{equation}
The server collects all the updates from the clients and performs federated aggregation. In the case of standard FedAvg, the server updates the model with the average of the gradients.
\begin{equation}
    P^r_{(t, s)} = P^{r - 1}_{(t, s)} + \sum_{c \in S^r_t}(\Delta P^r_{(t, c)}).
\end{equation}\label{eq:fed_aggregation}
This is equivalent to applying SGD with {pseudo-gradient} $\sum_{c \in S^r_t}(\Delta P^r_{(t, c)})$ with a learning rate $\eta = 1$ \cite{FedOpt}.
In a more general form, the server update can be written as:
\begin{equation}
    P^r_{(t, s)} = P^{r - 1}_{(t, s)} + \eta^r_{(t)}\Delta_t^r\label{eq:fed_opt_aggregation}
\end{equation}
where $\eta^r_{(t)}$ is the server learning rate at round $r$ for task $t$, and $\Delta^r_t$ is the pseudo-gradient which is $\sum_{c \in S^r_t}(\Delta P^r_{(t, c)})$ in case of FEDSGD. With the notion of momentum with a decay parameter $\beta_1$, the pseudo gradient can be expressed as $\Delta^r_t = \beta_1\Delta^{r - 1}_t + (1 - \beta_1) \sum_{c \in S^r_t}(\Delta P^r_{(t, c)})$. 
The federated aggregation can be further generalized to more adaptive gradient descent algorithms. This family of algorithms is described as FedOpt in \cite{FedOpt}. 
The update rule in FedOpt is expressed as,
\begin{equation}
    P^r_{(t, s)} = P^{r - 1}_{(t, s)} + \eta^r_{(t)} \frac{\Delta_t^r}{\sqrt{v_t^r} + \tau} \label{eq:fed_opt_agg_gen}
\end{equation}
where $v_t^r$ is the exponential average of the square of the pseudo-gradient (second moment) that captures the variance. This is calculated differently in each of the variants of FedOpt as summarized in Table \ref{tab:adaptive_fed_equations}.
The parameter $\tau$ controls the degree of adaptiveness of the algorithm with smaller $\tau$ implying more adaptive. FedOpt keeps track of the history of gradients over several rounds and updates the model in a smooth fashion, it results in avoiding abrupt changes to the global model thereby reducing client drift. 

\begin{table}[t]
\vspace{-3mm}
  \centering
  \caption{Calculation of second order momentum of pseudo-gradients in different adaptive federated optimizers \cite{FedOpt}.}
  \label{tab:adaptive_fed_equations}
  \begin{adjustbox}{max width=0.6\columnwidth}
  \begin{tabular}{@{}lc@{}}
    \toprule
    Method & Update Rule for Second Moment\\
    \midrule
    FEDADAGRAD & $v_t^r = v_t^{r - 1} + {\Delta_t^r}^2 $ \\
    FEDYOGI & $v_t^r = v^{r-1}_t - (1-\beta_2) {\Delta_t^r}^2 \text{sign}(v_t^{r-1} - {\Delta_t^r}^2)$  \\
    FEDADAM &$v_t^r = \beta_2 v_t^{r - 1} + (1 - \beta_2){\Delta_t^r}^2$ \\
    \bottomrule
  \end{tabular}
 \end{adjustbox}
  
  \vspace{-5mm}
\end{table}

\begin{figure}[t]
\vspace{-6mm}
\begin{algorithm}[H]
\caption{FCL Training Method. \colorbox{gray!30}{Case 1} \colorbox{blue!30}{Case 2}, \colorbox{orange!30}{Case 3}, }\label{alg:combined}
\hspace*{\algorithmicindent} \textbf{Input:} Clients $c \in [N]$ Tasks $t \in [T]$, No. of rounds $R$, Local datasets $D_c \forall c \in [N]$  \\
\hspace*{\algorithmicindent} \textbf{Output:} Trained task-specific blocks $P_{(t, s)} \forall t \in [T]$
\begin{algorithmic}[1]
\For{client $c = 1, ... N$}
    \State \colorbox{orange!30}{$task\_order_c$ = permute(1...T)}
\EndFor
\For{round $r = 1, ... R$}
    \State $Q = R/T$
    \For{client $c = 1, ... N$ }
        \State \colorbox{gray!30}{task $t \sim [T]$ sampled uniformly at random}
        \State \colorbox{blue!30}{task $t = \lceil{r//Q}\rceil$}
        \State \colorbox{orange!30}{task $t = task\_order_c[\lceil{r//Q}\rceil]$}
        \State Local update for $K$ epochs on $P^{r}_{(t, c)}$ (Eq. )
        \State $S_t^r = S_t^r  \cup c$
    \EndFor
    \For{task $t = 1, ... T$ }
        \If{$S_t^r \ne \phi$}
            \State Global update on task $t$, $P^r_{(t, s)}$
        \EndIf
    \EndFor
\EndFor
\end{algorithmic}
\end{algorithm}
\vspace{-15mm}
\end{figure}

We define a phase (Q) as the number of rounds a client works on each task and break the overall training into $T$ phases. In both cases 2 and 3, we change the task at each client only after Q rounds. For instance, in the example shown in Fig. \ref{fig:task_order_revised}, the total rounds (R) is 16 and the number of tasks (T) is 4 resulting in Q = 4. Hence, the task is changed in all the clients after every 4 rounds.
In cases 2 and 3, we divide R rounds of federated training into T phases of Q rounds each. Then, we perform federated aggregation on the updates from all the clients with one task every Q rounds. This ensures that each client performs an equal amount of training in terms number of local epochs for all three cases resulting in a fair comparison. 
Algorithm \ref{alg:combined} provides a flow of the FCL for different task ordering.
We focus mainly on the case of shuffled task order without repetition depicting asynchronous FCL (Case 3) to perform extensive analysis and use cases 1 and 2 for comparison.




\section{Experiments}

We conduct extensive experiments spanning different aspects of federated learning as well as continual learning on three benchmark datasets --- CIFAR100 \cite{CIFAR,CL_SI}, MiniImagenet \cite{miniImagenet} and Decathlon Challenge for Continual Learning \cite{decathlon}. CIFAR100 is a standard benchmark dataset widely used in computer vision containing 50000 training images and 10000 test images with 32x32 pixel dimensions and 3 channels. We split the training data into 10 tasks of 10 classes each. 
MiniImagenet is a 100 class dataset sampled from the ImageNet data. The resolution of the images is reduced to 84x84. Similar to CIFAR100, we divide the miniImagenet into 10 tasks of 10 classes each. Decathlon benchmark is a collection of ten well-known datasets used to benchmark continual learning methods on visual domain. Each dataset in the collection is defined as a separate task.
Across all three datasets, the test set is held out and all the reported evaluation is performed on the entire test set. 
We divide the datasets across the clients uniformly at random to obtain identically distributed data (IID) and use Dirichlet distribution to simulate non-IID scenarios. We use the IID case across our experiments to identify the effect of varying each parameter independently and study the impact of non-IID distribution in section \ref{section:ablation_study}.
We use standard ResNet18 \cite{he2016deep} as the backbone model across all our experiments.
{
We examine the behavior of client drift in different ordering of tasks and show the extent of the problem in \ref{section:task_order}.
While we use all three datasets mentioned above for showing the effectiveness of our framework in section \ref{section:different_datasets}, we use CIFAR100 as the candidate dataset for comparing with related work in section \ref{section:comp_prev_work}. 
We use the set of hyperparameters shown in Table \ref{tab:parameter_table} as default values and perform a principled ablation study by varying each of the parameters in the section \ref{section:ablation_study}.  }

\begin{table}[t]
\vspace{-3mm}
  \centering
  \caption{Default hyperparameters used across all the experiments unless specified otherwise.}
  \label{tab:parameter_table}
  \begin{adjustbox}{max width=0.85\columnwidth}
  \begin{tabular}{@{}ll|ll@{}}
    \toprule
    Parameter & Value & Parameter & Value \\
    \midrule
    Dataset & CIFAR100 & Client Optimizer & SGD \\
    Backbone Model & ResNet18 & Client LR ($\mu$) & 0.05 \\
    No. of Tasks (T) & 10 & No. of Local Epochs (K) & 2 \\
    No. of Clients (N) & 5 & Data Distribution & IID \\
    No. of Rounds (R) & 300 & Server Optimizer & FEDADAM \\
    No. of Rounds per phase (Q=R/T) & 30 & Server LR ($\eta$) & 0.5 \\
    \bottomrule
  \end{tabular}
  \end{adjustbox}
  
  \vspace{-2mm}
\end{table}


\noindent
\textbf{Evaluation Metrics:} The key performance metric is the average accuracy of the model on all tasks.
Additionally, we also keep track of a metric to measure the client drift. Similar to backward transfer in continual learning \cite{lopez2017gradient} which measures the effect of learning a new task $t$ on the performance of previous tasks $ t' < t$,
we define federated negative backward transfer $BWT_f$ to quantify the client drift introduced by the latter clients on the learned model. We capture the test accuracy of all tasks at each federated round resulting in a accuracy matrix $A \in \mathbb{R}^{TxR}$ and define the evaluation metrics of average accuracy $(ACC = \frac{1}{T}\sum_{t = 1}^{T} A_{t, R})$ and $BWT_f$ as follows:
\begin{equation}
\begin{aligned}
BWT_f = \frac{1}{T^2}\sum_{t = 1}^{T}\sum_{p = 1}^{T} (A_{t, pQ} - A_{t, pQ-1}). 
\end{aligned}
\end{equation}
Here, Q is the number of rounds per phase. With $BWT_f$, we are capturing the drop in accuracy resulting from new clients updating the model after every phase and take the average over all the tasks. 

\begin{figure}[t]
\centering

\subfloat[The model accuracy on each of the tasks in different order of tasks.]{%
  \includegraphics[clip,width=0.45\textwidth]{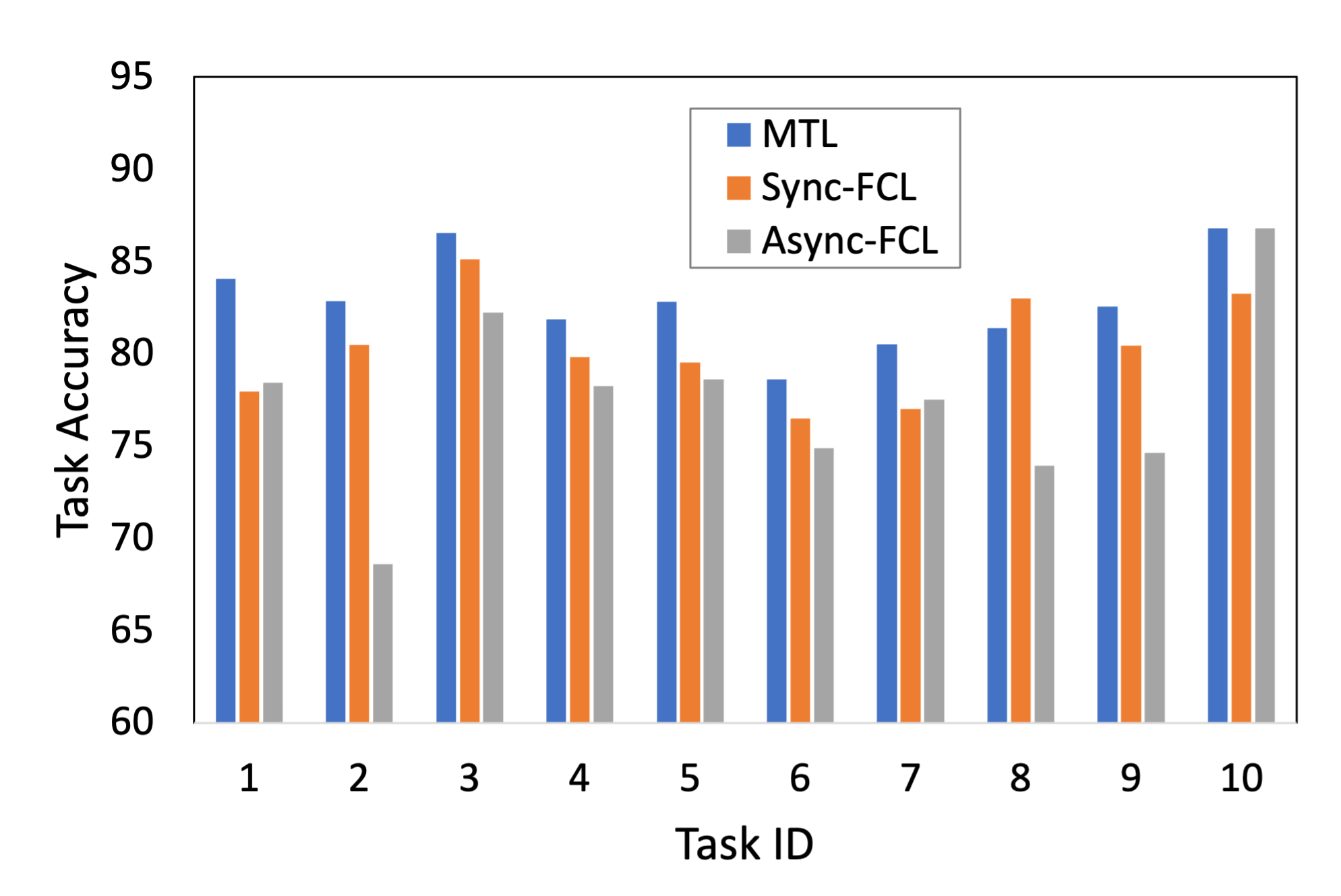}%
  \label{fig:task_order_acc_comp}
}
\qquad
\subfloat[Client drift in each of the three order of tasks.]{%
  \includegraphics[clip,width=0.45\textwidth]{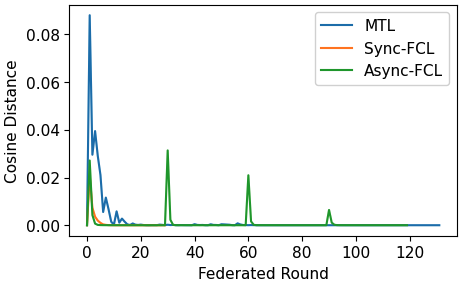}%
  \label{fig:client_drift_cosine_comp}
}

\caption{Illustration of client drift in asynchronous FCL. (a) Comparison of the different ordering of tasks. Note the performance drop in asynchronous FCL (Case 3) compared to synchronous FCL (Case 2) in most of the tasks, while FMTL performs the best among the three cases.  (b) Visualization of client drift in terms of cosine distances between the models from consecutive rounds for the three cases of the relative order of tasks.  }
\vspace{-5mm}
\end{figure}

\vspace{-3mm}
\subsection{Client Drift in Different Order of Tasks}\label{section:task_order} \vspace{-2mm}
Here we compare the different order of tasks described in section \ref{section:framework}.
We present the task-wise accuracy for all three cases on the CIFAR100 dataset in Fig. \ref{fig:task_order_acc_comp}.
Since the updates from different clients is repeated in FMTL, the overwriting is avoided resulting in the highest performance among the three cases (i.e. baseline average accuracy of 82.79\%).
In synchronous FCL, since all the clients are synchronously training on the same task, the client drift is at a minimum hence the accuracy is high. 
However, in the case of asynchronous FCL, where the tasks are shuffled and there is no repetition, updates from the later rounds can overwrite the information learned from the clients in earlier rounds resulting in a significant performance drop. 
We use cosine distance between the models following eq. \ref{eq:client-drift} to quantify and visualize client drift. We take the weights of the fully connected layers for each task as the proxy for the model and calculate the cosine distance between the models from consecutive rounds 
and plot for all three cases in Fig.\ref{fig:client_drift_cosine_comp}. Note the spikes in cosine distance in the case of asynchronous FCL indicates that the clients in the later rounds overwrite the information learned from the earlier rounds highlighting the existence of the problem of aforementioned client drift in federated continual learning systems. In contrast, the case of FMTL shows significant deviation only during the initial stages of learning and becomes stable once the model is sufficiently trained.{ We show that using adaptive optimizers can reduce the impact of client drift and present our findings in the next section.
}

\vspace{-3mm} \subsection{Adaptive Optimizers and Client Drift}\label{section:different_datasets} \vspace{-2mm}
\begin{table*}[t]
\vspace{-3mm}
\centering
\caption{Summary of performance of different variants of adaptive optimizers on three different datasets. We highlight the top two optimizers achieving high accuracy and and low $BWT_f$ for each dataset. Adaptive optimizers consistently perform well both in achieving high accuracy and reducing the $BWT_f$.}
  \label{tab:cifar100_results}
\begin{adjustbox}{max width=0.9\textwidth}
\begin{tabular}{l|l|ccc|ccc}
\toprule
\multirow{2}{*}{Dataset} & \multirow{2}{*}{Optimizer}     & \multicolumn{3}{c}{$ACC$} & \multicolumn{3}{c}{$BWT_f$} \\
 &  & $\eta$ = 1.0      & $\eta$ = 0.5    & $\eta$ = 0.1   & $\eta$ = 1.0      & $\eta$ = 0.5    & $\eta$ = 0.1   \\
  \midrule
\multirow{4}{*}{CIFAR100} & FEDSGD     & 77.59  & 79.27  & 78.51 & -5.76  & -1.04  & -0.63 \\
& FEDADAM    & 77.85  & \textbf{80.04}  & 77.98 & -3.61  & -0.84  & -0.55 \\
& FEDADAGRAD & 77.55  & 78.33  & 78.98 & -4.42  & -1.39  & {-0.51} \\
& FEDYOGI    & 76.81  & {79.33}  & 78.97 & -4.44  & -1.77  & \textbf{-0.50} \\
  \midrule
\multirow{4}{*}{MiniImageNet} & FEDSGD                     & 56.01                 & 56.82                   & 52.44                   & -1.86                 & -0.45                   & -0.12                   \\
& FEDADAM                    & 56.89                 & {57.28}                   & 53.73                   & -1.75                 & -0.16                   & \textbf{-0.08}                   \\
& FEDADAGRAD                 & 56.81                 & 57.14                   & 52.23                   & -1.72                 & -0.23                   & {-0.08}                   \\
& FEDYOGI                    & 56.75                 & \textbf{57.34}                   & 52.98                   & -1.98                 & -0.21                   & -0.11 \\
  \midrule
\multirow{4}{*}{Decathlon Challenge}
& FEDSGD                     & 73.44                 & 74.62                   & 74.01                   & -3.45                 & -1.42                   & -0.84                   \\
& FEDADAM                    & 73.24                 & {74.74}                   & 74.52                   & -2.82                 & -1.04                   & -0.67                   \\
& FEDADAGRAD                 & 73.82                 & \textbf{75.03}                   & 74.61                   & -3.23                 & -1.45                   & {-0.53}                   \\
& FEDYOGI                    & 72.56                 & 74.38                   & 73.63                   & -3.16                 & -1.00                   &  \textbf{-0.46}                   \\
\bottomrule
\end{tabular}
\end{adjustbox}

  \vspace{-5mm}
\end{table*}

In this section, we highlight the advantage of using adaptive optimizers in federated aggregation to avoid the aforementioned problem of client drift. We compare the variants of adaptive optimizers with three different learning rates of 1.0, 0.5, and 0.1 on three datasets --- CIFAR100, MiniImageNet, and Decathlon Challenge (shown in Table \ref{tab:cifar100_results}). In general, we observe a trend of adaptive optimizers reducing the client drift as reflected with low negative $BWT_f$. At the same time, the final accuracy reached is higher than the baseline model that uses FedAvg (i.e., FEDSGD with a learning rate of 1.0). For example, FEDSGD with a learning rate of 1.0, which is equivalent to standard FedAvg, suffers from a severe client drift resulting in a 5.76\% drop in accuracy for CIFAR100. In contrast, FEDADAM reduces the $BWT_f$ to 3.61\% for the same setting. 

\begin{figure}[t]
\centering

\subfloat[Snapshot of training progress.]{%
  \includegraphics[clip,width=0.45\textwidth]{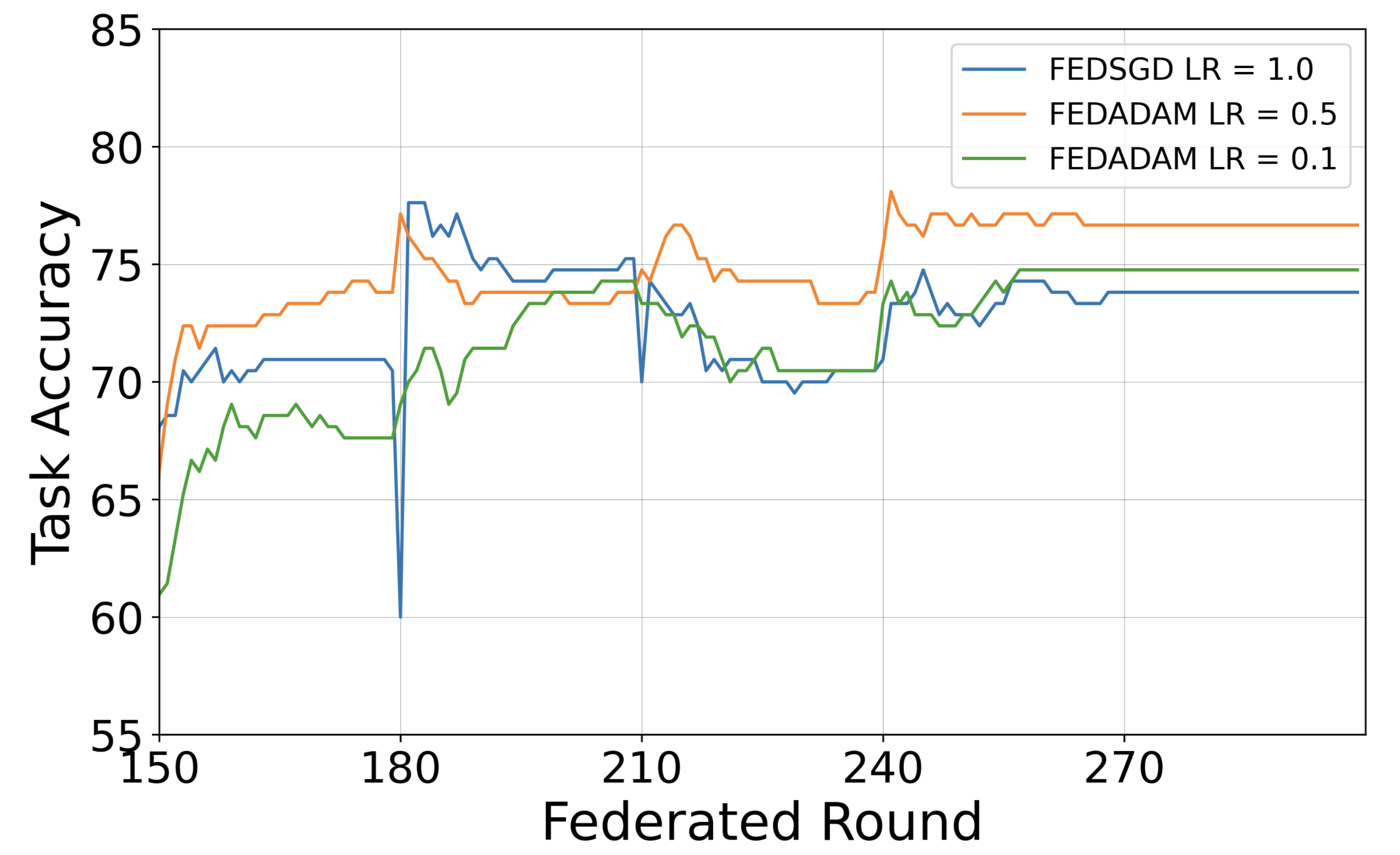}%
  \label{fig:zoom_in}
}
\qquad
\subfloat[Visualization of client drift.]{%
  \includegraphics[clip,width=0.45\textwidth]{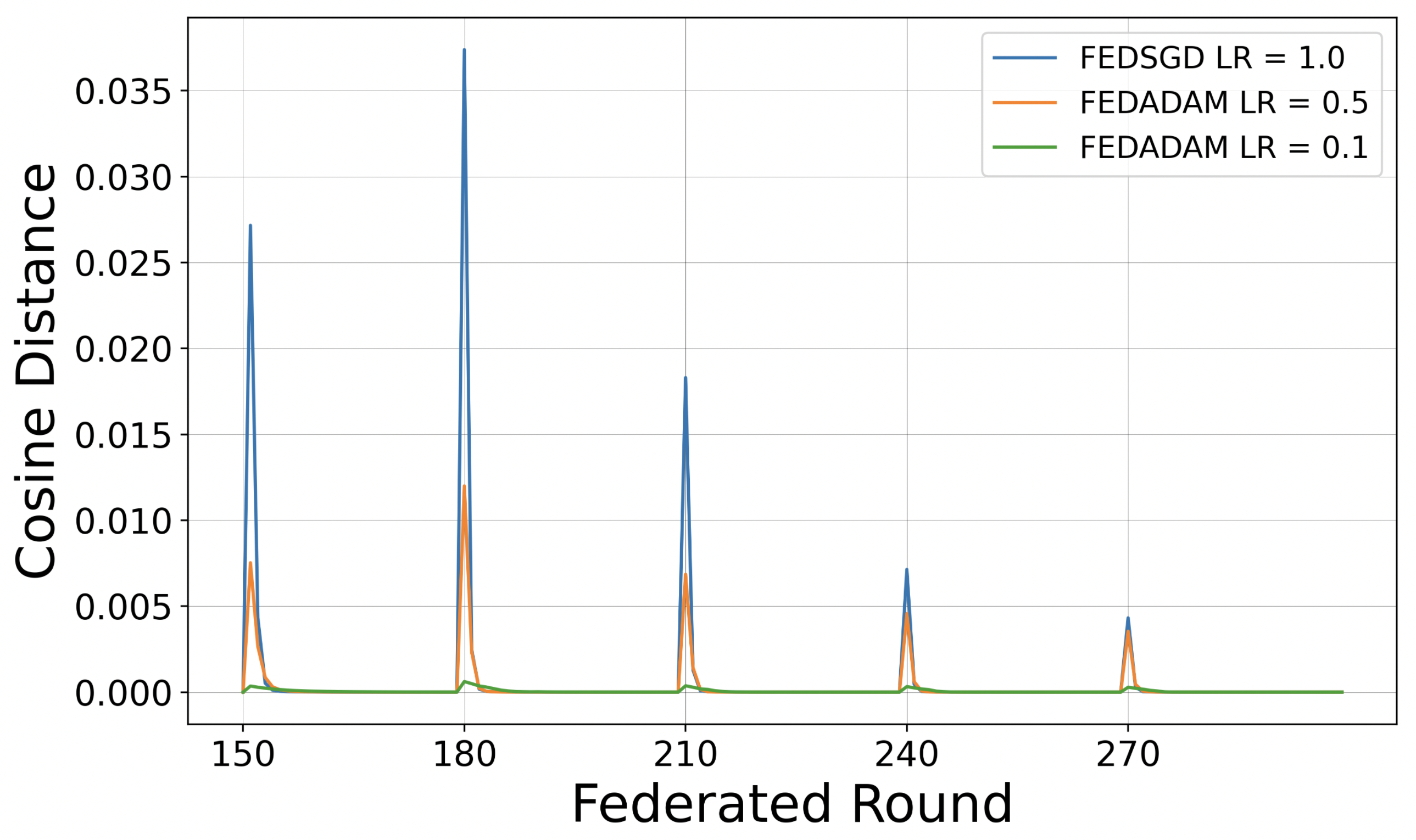}%
  \label{fig:cosine_dist}
}

\caption{Visualization of training progress of FEDADAM as compared to FEDSGD on one of the tasks of CIFAR100. (a) Shows the trend of model accuracy on one of the tasks over the course of training demonstrating the effectiveness of the FEDADAM optimizer in minimizing the sudden accuracy drops seen in FEDSGD. (b) The steep spikes in client drift measured in terms of cosine distances between the models from consecutive rounds illustrate the advantage of using FEDADAM over FEDSGD. }
\vspace{-5mm}
\end{figure}


{
To further understand the impact of client drift and the effectiveness of using an adaptive optimizer we zoom in on the training progress of Task 1 on the CIFAR100 dataset across federated rounds in Fig. \ref{fig:zoom_in}. Note the sharp drops in accuracy at the boundaries of phases (at rounds 180 and 210) with FEDSGD which indicate that the updates from the current rounds are overwriting the model learned using the data from clients in the previous rounds. 
We compare the cosine distances for FEDADAM with learning rates 0.5 and 0.1 to the baseline FEDSGD with a learning rate of 1.0. 
FEDADAM avoids sudden jumps in the model updates by using the history of gradients to make the updates smooth. At the same time, if we use a lower learning rate, the updates themselves become small, hence, the convergence rate of the model is adversely affected. 
We expand on this further by studying the behavior of the training method for different learning rates and the number of rounds in section \ref{section:ablation_study}. 
}

\vspace{-3mm} \subsection{Comparison to Previous Work}\label{section:comp_prev_work} \vspace{-2mm}
{
We compare our aynchronous FCL results with FEDADAM training against previous work of FedWeIT \cite{fedweit} on CIFAR100 dataset in Table \ref{tab:comp_fedweit}.
Note that while the dataset partition is different since the reported accuracy is the average across all tasks for CIFAR100, the comparison is fair. Our method outperforms FedWeIT by a significant margin in terms of accuracy while having <1\% client drift. The improved performance can be attributed to the separate task parameters of the NetTailor architecture thereby eliminating the catastrophic forgetting due to interference between the tasks. Also, the FEDADAM optmization helps in reducing the overall client drift. Having separate parameters for each task aids in scaling our framework to more tasks as we can add an unlimited number of tasks without interfering with each other whereas it is uncertain how FedWeIT will scale to additional tasks. Additionally, we compare the model size and average communication cost per round for sharing the gradients from clients to server (C2S) and broadcasting the updated model to the clients (S2C) for a 5 client system trained on 10 tasks. While our model size is higher compared to FedWeIT, the communication cost per round is significantly lower since we are only communicating the gradients of the task-specific modules. This can be further reduced by applying gradient compression methods \cite{lin2017deep,albasyoni2020optimal}.}

\begin{table}[t]
\vspace{-3mm}
  \centering
    \caption{Comparison to previous work in terms of accuracy, forgetting, model size and communication cost on CIFAR100 dataset. C2S and S2C denote communication cost from client to server and server to client respectively.}
  \label{tab:comp_fedweit}
  \begin{adjustbox}{max width=0.6\columnwidth}
  \begin{threeparttable}
  \begin{tabular}{@{}l|c|c|c|c@{}}
    \toprule
    Method & Accuracy & $BWT_f$ & Model & Comm. cost per \\
    &&&Size& round (C2S + S2C)\\
    \midrule
    FedWeIT\cite{fedweit} & 55.16\% & - & 75{\small MB} & 18.5{\small MB} + 53.5{\small MB}\\
    Ours & 80.04\% & 0.84\% & \, 90MB\tnote{*} \,  & 4.7MB + 4.7MB \\
    \bottomrule
  \end{tabular}
  \begin{tablenotes}
  \item[*] 43MB (backbone) + 10 tasks $\times$ 4.7MB (per task)
  \end{tablenotes}
 
 \end{threeparttable}
 \end{adjustbox}

\end{table}

\vspace{-3mm} \subsection{Ablation Studies}\label{section:ablation_study} \vspace{-2mm}


\begin{figure*}[t]
  \begin{center}
    \includegraphics[width=\textwidth]{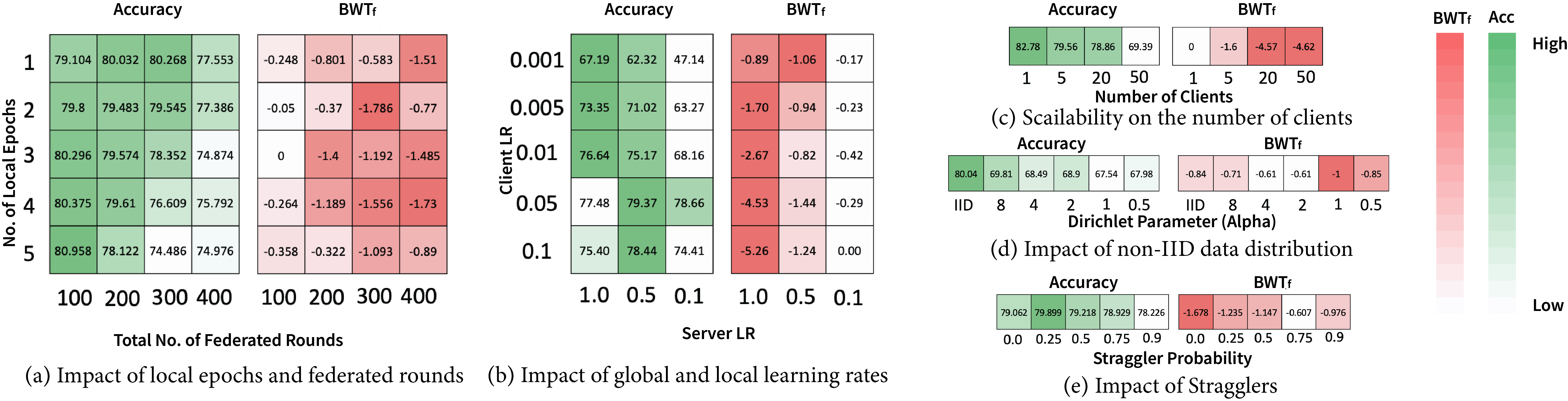}
  \end{center}
  \vspace{-3mm}
  \caption{
    Ablation study of different hyperparameters involved. We report the performance of the system in terms of ACC and $BWT_f$. The heatmaps are color-coded with green for higher accuracy (higher is better) and red for larger client drift (lower magnitude is better). (a) The trend of Accuracy and $BWT_f$ with local and global learning rates. 
(b) Trend of accuracy and $BWT_f$ with the number of rounds and local epochs. (c) Performance trend with varying the number of clients. (d) Non-IID Trend. (e) Performance trend in the presence of stragglers.
  }
   \label{fig:ablation_full}
   \vspace{-6mm}
\end{figure*}

\noindent

\textbf{Number of rounds and local epochs:} { The number of local epochs and the total number of federated rounds are important factors in achieving convergence in federated learning systems. Here, we vary the number of local epochs from 1 to 5 for a different number of federated rounds ranging from 100 to 400. As summarized in Fig. \ref{fig:ablation_full}a, we observe that having a large number of federated rounds reduces the accuracy as client drift accumulates over several rounds.}

\noindent
\textbf{Server and client learning rates:} Here we vary both client and server learning rates and measure the final performance of the system (see Fig. \ref{fig:ablation_full}b). The learning rate at the client dictates how effectively the local models are being learned. This follows the trend of standard deep learning where using a lower learning rate will make the convergence slower but increasing the learning rate after some point makes the training unstable resulting in a diverging model. Therefore, we see an improvement in average accuracy as we increase the client learning rate up to 0.05 but the accuracy drops as we go to 0.1. { Similarly, decreasing the server learning rate makes the federated aggregation process smooth, resulting in improved performance both in terms of accuracy and client drift. However, reducing it further makes the updates extremely small resulting in slower convergence. Hence, for a fixed number of rounds, the final model has lower accuracy.}

\noindent
\textbf{Scalability:} In this experiment (Fig. \ref{fig:ablation_full}c), we vary the number of clients to observe the final accuracy reached by the models. Naturally, as the number of clients increases, there are more opportunities for newer clients to overwrite the learned model. Hence the performance of the system is reduced as the number of clients increases. 

\noindent
\textbf{Non-IID data:} Since practical federated learning systems often have a non-IID data distribution among the clients, we simulate the skewed distribution (Fig. \ref{fig:ablation_full}d) by sampling the proportions of each class based on the Dirichlet distribution following previous works \cite{FedMA,yurochkin2019bayesian}. By varying the parameter (alpha) of the Dirichlet distribution, we can control the skewness of the data distribution. Here, having a low alpha indicates that the data distribution is more skewed. As we vary the alpha value progressively from 8 to 0.5, we observe a decrease in performance both in terms of accuracy as well as $BWT_f$. This implies that the skewness in the system amplifies the client drift.

\noindent
\textbf{Stragglers:} Finally, it is common to have heterogeneous clients with different capabilities in terms of compute, memory and communication bandwidth. Hence, it is not always feasible to assume all the clients communicate their gradients in time to participate in the federated aggregation process. We measure the robustness of the system with respect to stragglers by varying the probability of the clients failing to communicate their gradients to the server. We progressively increase the probability of each client showing a straggler behavior (Fig. \ref{fig:ablation_full}e). We observe that while the performance goes down it is still within less than a 2\% drop from the baseline case, where, all the clients are guaranteed to communicate the updates. This shows the robustness of the system to stragglers.

\section{Conclusion and Limitations}
In this work, we present a framework for federated continual learning and identify client drift as the central challenge. We show that using adaptive federated optimizers can stabilize the federated aggregation and hence reduce the client drift. We provide a comprehensive analysis of different hyperparameters impacting the performance of the system. 

{
\textbf{Limitations:} While we propose a framework to perform federated continual learning with reduced client drift, there are numerous unsolved challenges for achieving a scalable and robust federated continual learning system. For instance, while we reduce client drift with adaptive optimizers, the problem is not fully solved. We observe in our ablation studies that as we increase the number of rounds, the clients in the later rounds eventually overwrite the model thereby reducing the accuracy of the final model. }Additionally, while NetTailor based approach removes catastrophic forgetting, the model size scales linearly with number of tasks making it impractical when we have large number of tasks. It is necessary to examine other continual learning methods that use a single model across all tasks such as Gradient Projection Memory (GPM) \cite{saha2021gradient} and Relevance Mapping Networks (RMN) \cite{kaushik2021understanding}. This will involve two orthogonal challenges of avoiding overwriting the model between tasks as well as between clients and additional scalability challenges which we leave for future work.
\clearpage
%
%
\bibliographystyle{splncs04}
\bibliography{egbib}
\end{document}